\documentclass[journal=jcisd8,manuscript=article]{achemso}

\usepackage[english]{babel}
\usepackage[utf8x]{inputenc}
\usepackage{chemformula}
\usepackage[T1]{fontenc}
\usepackage{amsmath,amsfonts}
\usepackage{graphicx}
\usepackage[colorlinks=true, allcolors=blue]{hyperref}
\usepackage{booktabs}
\usepackage{makecell,pbox}
\setkeys{acs}{
	maxauthors=100
}

\author{Seokho Kang}
\email{s.kang@skku.edu}
\phone{+82 31 290 7596}
\fax{+82 31 290 7610}
\affiliation[SKKU]
{Department of Systems Management Engineering, Sungkyunkwan University, 2066 Seobu-ro, Jangan-gu, Suwon 16419, Republic of Korea}

\author{Kyunghyun Cho}
\affiliation[NYU]{Department of Computer Science \& Center for Data Science, New York University, 60 5th Avenue, New York, NY 10011, USA}
\alsoaffiliation[FAIR]{Facebook AI Research, 770 Broadway, New York, NY 10003, USA}
\alsoaffiliation[CIFAR]{CIFAR Azrieli Global Scholar, Canadian Institute for Advanced Research, 661 University Avenue, Toronto, ON M5G 1M1, Canada}

\title{Conditional Molecular Design with Deep Generative Models}

\begin{document}

\begin{abstract}
Although machine learning has been successfully used to propose novel molecules that satisfy desired properties, it is still challenging to explore a large chemical space efficiently. In this paper, we present a conditional molecular design method that facilitates generating new molecules with desired properties. The proposed model, which simultaneously performs both property prediction and molecule generation, is built as a semi-supervised variational autoencoder trained on a set of existing molecules with only a partial annotation. We generate new molecules with desired properties by sampling from the generative distribution estimated by the model. We demonstrate the effectiveness of the proposed model by evaluating it on drug-like molecules. The model improves the performance of property prediction by exploiting unlabeled molecules, and efficiently generates novel molecules fulfilling various target conditions.
\end{abstract}

\section{Introduction}

The primary goal of molecular design is to propose novel molecules that satisfy desired properties, which has been challenging due to the difficulty in efficiently exploring a large chemical space. In the past, molecular design has been largely driven by human experts. They would suggest candidate molecules that are then evaluated through computer simulations and subsequent experimental syntheses.\cite{quantum01} This is time-consuming and costly, and is inadequate when many candidate molecules must be considered. In recent decades, machine learning based approaches have been actively studied as efficient alternatives to expedite the molecular design processes.\cite{pred01,pred02,pred03}

\begin{figure}[!t]
    \centering
    \includegraphics[width=0.9\textwidth]{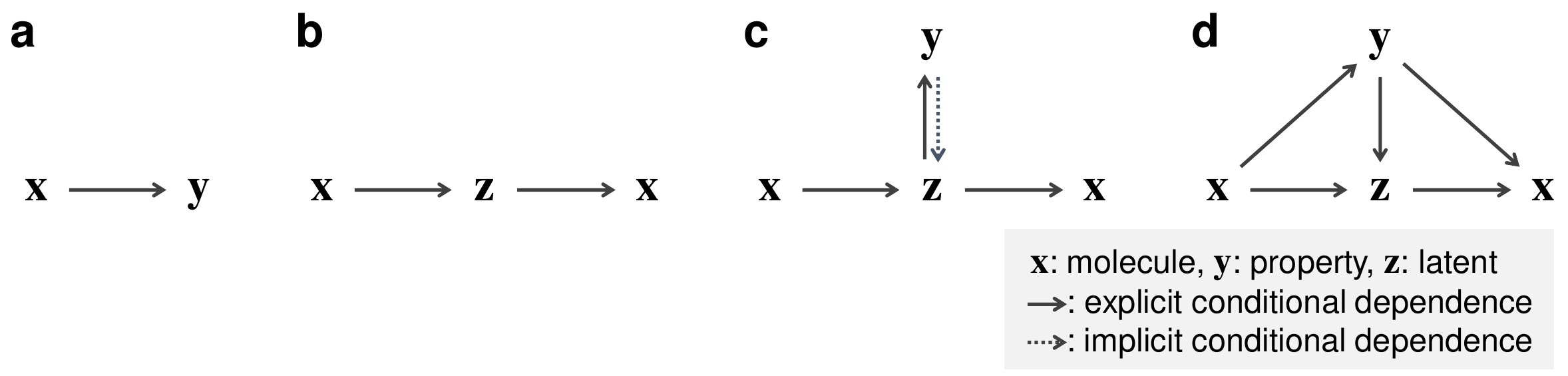}
    \caption{Schematic diagram of machine learning applications to molecular design: \textbf{(a)} property prediction, \textbf{(b)} molecule generation, \textbf{(c)} conditional molecular design (previous work), and \textbf{(d)} conditional molecular design (present work).}
    \label{figure:1}
\end{figure}

A conventional approach is to build a prediction model that estimates the properties of a given molecule, as shown in \autoref{figure:1}\textbf{(a)}. Molecules with desired properties are then chosen after screening a possible set of candidate molecules by this prediction model. \cite{hts01} The candidate molecules for screening need to be manually obtained from such sources as combinatorial enumerations of possible fragments \cite{comb01,comb02,comb03} and public databases.\cite{db01,db02} It is necessary to secure a sufficient amount of molecules that are properly labeled with properties, as prediction accuracy typically depends on the number of labeled molecules and the quality of the labels. Early work has attempted to transform a molecule into a hand-engineered feature representation, so-called molecular fingerprint, and to use it as an input to predict the properties.\cite{pred04, FP01, FP02} With recent advances of deep learning,\cite{deep01} prediction quality has improved by employing deep neural networks.\cite{deep02,pred05} Moreover, various recent studies have extracted features directly from graph representations of molecules to better predict the properties.\cite{conv01,conv02,conv03,cg01,cg02,cg03}

Another approach aims to automatically generate new molecules by building a molecule generation model, as in \autoref{figure:1}\textbf{(b)}. This approach learns to map a molecule on a latent space. From this space, it randomly generates new molecules that are analogous to those in the original set. Molecules randomly generated by this model can be used as candidates for screening with a separate property prediction model. Most existing studies on this approach have represented a molecule as a simplified molecular-input line-entry system (SMILES)\cite{smiles01} with a recurrent neural network (RNN). They include RNN language models,\cite{rnn01, rnn02,rnn03} variational autoencoders (VAE),\cite{vae01,rnn04,rnn05,rnn06} and generative adversarial networks (GAN)\cite{gan01,rnn07} with RNN decoders. More recently, the models directly generating the graph structures of molecules have also been proposed.\cite{graph01,graph02,graph03,graph04}

The molecule generation approach has been extended to conditional molecular design, generating a new molecule whose properties are close to a predetermined target condition. This is often done by finding a latent representation that closely reflects the target condition using a property prediction model, as in \autoref{figure:1}\textbf{(c)}. Previous work has proposed to use recursive fine tuning,\cite{rnn01} Bayesian optimization,\cite{rnn04} and reinforcement learning.\cite{rnn02,rnn03, rnn07} These methods generate molecules with intended properties not directly but by an additional optimization procedure often in the latent space. This is inefficient especially when multiple target conditions are considered.

Here we present a novel approach to efficiently and accurately generating new molecules satisfying designated properties. We build a conditional molecular design model that simultaneously performs both property prediction and molecule generation, as illustrated in \autoref{figure:1}\textbf{(d)}, using a semi-supervised variational autoencoder (SSVAE).\cite{vae02} Given a set of specific properties, conditional molecular design is done by directly sampling new molecules from a conditional generative distribution without any extra optimization procedure. The semi-supervised model can effectively exploit unlabeled molecules. This is advantageous particularly when only a small portion of molecules in the data are labeled with their properties, which is usual due to the expensive cost of labeling molecules.

\section{Methods}

\subsection{Model Architecture}

We adapt the original SSVAE\cite{vae02} to incorporate continuous output variables. SSVAE is a directed probabilistic graphical model that captures the data distribution in a semi-supervised manner. In the generative process of the SSVAE model, the input variable $\mathbf{x}$ is generated from a generative distribution $p_\theta (\mathbf{x}|\mathbf{y},\mathbf{z})$, which is parameterized by $\theta$ conditioned on the output variable $\mathbf{y}$ and latent variable $\mathbf{z}$. $\mathbf{y}$ is treated as an additional latent variable when $\mathbf{x}$ is not labeled, which necessitates introducing the distribution over $\mathbf{y}$. The prior distributions over $\mathbf{y}$ and $\mathbf{z}$ are assumed to be $p(\mathbf{y}) = \mathcal{N}(\mathbf{y}|\boldsymbol{\mu}_\mathbf{y}, \boldsymbol{\Sigma}_\mathbf{y})$ and $p(\mathbf{z}) = \mathcal{N}(\mathbf{z}|\mathbf{0},\mathbf{I})$. We use variational inference to address the intractability of the exact posterior inference of the model. We approximate the posterior distributions over $\mathbf{y}$ and $\mathbf{z}$ by $q_\phi (\mathbf{y}|\mathbf{x}) = \mathcal{N} (\mathbf{y}| \boldsymbol{\mu}_\phi (\mathbf{x}), \text{diag}(\boldsymbol{\sigma}^2_\phi (\mathbf{x})))$ and $q_\phi (\mathbf{z}|\mathbf{x},\mathbf{y}) = \mathcal{N}(\mathbf{z}| \boldsymbol{\mu}_\phi (\mathbf{x},\mathbf{y}), \text{diag}(\boldsymbol{\sigma}^2_\phi (\mathbf{x},\mathbf{y})))$, both of which are parameterized with $\phi$. For the semi-supervised learning scenario where some values of $\mathbf{y}$ are missing, the missing values are predicted by $q_\phi (\mathbf{y}|\mathbf{x})$.

In our framework for conditional molecular design, $\mathbf{x}$ and $\mathbf{y}$ denote a molecule and its continuous-valued properties, respectively. In this study, we consider molecules that can be represented by SMILES, which has been commonly used in the recent related work.\cite{rnn01,rnn02,rnn03,rnn04,rnn05,rnn07} SMILES encodes the graph structure of a molecule in a compact line notation by depth-first traversal into a sequence with a simple vocabulary and grammar rules.\cite{smiles01} For example, a benzene is described in the form of SMILES as \texttt{c1ccccc1}. A molecule representation $\mathbf{x}$ is then formed as a sequence of one-hot vectors describing a SMILES string $(\mathbf{x}^{(1)},\ldots,\mathbf{x}^{(j)},\ldots,\mathbf{x}^{(l)})$, where each one-hot vector $\mathbf{x}^{(j)}$ corresponds to the index of a symbol in a predefined vocabulary and $l$ is the length of the sequence. The vocabulary consists of all unique characters in the data, except for atoms represented by two characters (\emph{e.g.}, \texttt{Si}, \texttt{Cl}, \texttt{Br}, and \texttt{Sn}) which are considered single symbols. A vector $\mathbf{y}$ consists of $m$ scalar values $(y^1,\ldots,y^m)$, where $m$ is the number of properties of a molecule.

We use an RNN to model $p_\theta$ and $q_\phi$. The SSVAE model is composed of three RNNs, which are the predictor network $q_\phi (\mathbf{y}|\mathbf{x})$, the encoder network $q_\phi (\mathbf{z}|\mathbf{x},\mathbf{y})$, and the decoder network $p_\theta (\mathbf{x}|\mathbf{y},\mathbf{z})$. We use bidirectional RNNs \cite{birnn} for the predictor and encoder networks, while the decoder network is a unidirectional RNN. The input to the encoder network at each time step $j$ contains $\mathbf{x}^{(j)}$ and $\mathbf{y}$. The decoder network, which generates sequences, takes the output of the current time step $j$, $\mathbf{y}$, and $\mathbf{z}$ as the input at time $j$+1.

\subsection{Objective Functions}

We define two loss functions $\mathcal{L}(\mathbf{x},\mathbf{y})$ and $\mathcal{U}(\mathbf{x})$ corresponding to labeled and unlabeled instances, respectively. The variational lower bound $-\mathcal{L}(\mathbf{x},\mathbf{y})$ of the log-probability of a labeled instance $(\mathbf{x},\mathbf{y})$ is:
\begin{equation}
\begin{split}
\log p(\mathbf{x},\mathbf{y}) \geq & \mathbb{E}_{q_{\phi}{(\mathbf{z}|\mathbf{x},\mathbf{y})}} \left[ \log p_\theta (\mathbf{x}|\mathbf{y},\mathbf{z}) + \log p (\mathbf{y}) + \log p(\mathbf{z}) - \log q_\phi (\mathbf{z}|\mathbf{x},\mathbf{y}) \right]\\
= & \mathbb{E}_{q_{\phi}{(\mathbf{z}|\mathbf{x},\mathbf{y})}} \left[ \log  p_\theta (\mathbf{x}|\mathbf{y},\mathbf{z})\right] + \log p (\mathbf{y}) - \mathcal{D}_\text{KL} (q_\phi (\mathbf{z}|\mathbf{x},\mathbf{y}) ||  p(\mathbf{z}) )\\
= &- \mathcal{L}(\mathbf{x},\mathbf{y}). \\
\end{split}
\end{equation}
For an unlabeled instance $\mathbf{x}$, $\mathbf{y}$ is considered as a latent variable. The variational lower bound $-\mathcal{U}(\mathbf{x})$ is then:
\begin{equation}
\begin{split}
\log p(\mathbf{x}) \geq & \mathbb{E}_{q_{\phi}{(\mathbf{y},\mathbf{z}|\mathbf{x})}} \left[ \log p_\theta (\mathbf{x}|\mathbf{y},\mathbf{z}) + \log p (\mathbf{y}) + \log p(\mathbf{z}) - \log q_\phi (\mathbf{y},\mathbf{z}|\mathbf{x}) \right]\\
= & \mathbb{E}_{q_{\phi}{(\mathbf{y},\mathbf{z}|\mathbf{x})}} \left[ \log p_\theta (\mathbf{x}|\mathbf{y},\mathbf{z})\right]  - \mathcal{D}_\text{KL} (q_\phi (\mathbf{y}|\mathbf{x}) || p(\mathbf{y}) ) - \mathbb{E}_{q_{\phi}{(\mathbf{y}|\mathbf{x})}} \left[ \mathcal{D}_\text{KL} (q_\phi (\mathbf{z}|\mathbf{x},\mathbf{y})|| p(\mathbf{z}) ) \right] \\
=& - \mathcal{U}(\mathbf{x}),\\
\end{split}
\end{equation}
where $\log q_\phi (\mathbf{y},\mathbf{z}|\mathbf{x}) = \log q_\phi (\mathbf{y}|\mathbf{x})  + \log q_\phi (\mathbf{z}|\mathbf{x},\mathbf{y})$.

Given the data distributions of labeled $\widetilde{p_\text{l}}(\mathbf{x},\mathbf{y})$ and unlabeled cases $\widetilde{p_\text{u}}(\mathbf{x})$, the full loss function $\mathcal{J}$ for the entire dataset is defined as:
\begin{equation}\label{eq03}
\mathcal{J} = \sum_{(\mathbf{x},\mathbf{y}) \sim \widetilde{p_\text{l}}}{\mathcal{L}(\mathbf{x},\mathbf{y})} + \sum_{(\mathbf{x}) \sim \widetilde{p_\text{u}}}{\mathcal{U}(\mathbf{x})} + \beta \cdot \sum_{(\mathbf{x},\mathbf{y}) \sim \widetilde{p_\text{l}}} {    ||\mathbf{y}- \mathbb{E}_{q_\phi(\mathbf{y}|\mathbf{x})} \left[\mathbf{y}\right]||^2  },
\end{equation}
where the last term is mean squared error for supervised learning. The distribution $q_\phi (\mathbf{y}|\mathbf{x})$ is not estimated from the labeled cases without the last term, because $q_\phi (\mathbf{y}|\mathbf{x})$ does not contribute to $\mathcal{L}(\mathbf{x},\mathbf{y})$.\cite{vae02} The last term encourages $q_\phi (\mathbf{y}|\mathbf{x})$ to be predictive of the observed properties based on the labeled instances. As $\mathbf{y}$ is assumed to follow a normal distribution, $\mathbb{E}_{q_\phi(\mathbf{y}|\mathbf{x})} \left[\mathbf{y}\right]$ is equivalent to $\boldsymbol{\mu}_\phi (\mathbf{x})$. The hyper-parameter $\beta$ controls the trade-off between generative and supervised learning. It becomes fully generative learning when $\beta=0$, while it focuses more on supervised learning with a larger $\beta$.

\subsection{Property Prediction}
Once the SSVAE model is trained, property prediction is performed using the predictor network $q_\phi (\mathbf{y}|\mathbf{x})$. Given an unlabeled instance $\mathbf{x}$, the corresponding properties $\hat{\mathbf{y}}$ are predicted as below.
\begin{equation}
\hat{\mathbf{y}} \sim \mathcal{N}(\boldsymbol{\mu}_\phi (\mathbf{x}), \text{diag}(\boldsymbol{\sigma}^2_\phi (\mathbf{x})))
\end{equation}
The point estimate of $\hat{\mathbf{y}}$ can be obtained by maximizing the probability, which is equivalent to $\boldsymbol{\mu}_\phi (\mathbf{x})$.

\subsection{Molecule Generation}
We use the decoder network $p_\theta (\mathbf{x}|\mathbf{y},\mathbf{z})$ to generate a molecule. A molecule representation $\hat{\mathbf{x}}$ is obtained from $\mathbf{y}$ and $\mathbf{z}$ by
\begin{equation}
\hat{\mathbf{x}} = \underset{{\mathbf{x}}}{\arg\max}~{\log p_\theta (\mathbf{x}|\mathbf{y},\mathbf{z}}).
\end{equation}
At each time step $j$ of the decoder, the output $\mathbf{x}^{(j)}$ is predicted by conditioning on all the previous outputs $(\mathbf{x}^{(1)}, \ldots, \mathbf{x}^{(j-1)})$, $\mathbf{y}$, and $\mathbf{z}$, because we decompose $p_\theta {(\mathbf{x}|\mathbf{y},\mathbf{z})}$ as
\begin{equation}
p_\theta {(\mathbf{x}|\mathbf{y},\mathbf{z})}=\underset{j}{\prod}\text{ }{p_{\theta}(\mathbf{x}^{(j)}|\mathbf{x}^{(1)},\ldots,\mathbf{x}^{(j-1)}, \mathbf{y}, \mathbf{z})}.
\end{equation}
The optimal decoding solution $\hat{\mathbf{x}}$ can be obtained by maximizing the autoregressive distribution of $p_\theta {(\mathbf{x}|\mathbf{y},\mathbf{z})}$. This is however computationally intractable, because the search space grows exponentially with respect to the length of sequences. Sampling from the autoregressive distribution is simple and fast, but is vulnerable to the noise in sequence generation. Therefore, we use beam search to find an approximate solution efficiently, which has been successfully used to generate sequences with RNN.\cite{beam01,beam02,beam03} Beam search generates a sequence from left to right based on a breadth-first tree search mechanism. At each time step $j$, top-$K$ candidates are maintained. 

To generate an arbitrary molecule unconditionally, $\mathbf{y}$ and $\mathbf{z}$ are sampled from their prior distributions $p(\mathbf{y})$ and $p(\mathbf{z})$, respectively. For conditional molecular design given a target value for a property, $\mathbf{z}$ is sampled from $p(\mathbf{z})$, while the corresponding element of $\mathbf{y}$ is set to the target value and the other elements are sampled from the conditional prior distribution given the target value. For example, if we want to generate a new molecule whose first property is close to 0.5, the first element $y_1$ is set to 0.5 while the other elements are sampled from $p(y_2,\ldots{...},y_m | y_1=0.5)$.

\section{Results and Discussion}

\subsection{Dataset}
We collect 310,000 SMILES strings of drug-like molecules randomly sampled from the ZINC database.\cite{db01} We use 300,000 molecules for training and the remaining 10,000 molecules for testing the property prediction performance. The SMILES strings of the molecules are canonicalized using the RDKit package,\cite{rdkit} and then are transformed into sequences of symbols occurring in the training set. The vocabulary contains 35 different symbols including \{\texttt{1}, \texttt{2}, \texttt{3}, \texttt{4}, \texttt{5}, \texttt{6}, \texttt{7}, \texttt{8}, \texttt{9}, \texttt{+}, \texttt{-}, \texttt{=}, \texttt{$\#$}, \texttt{(}, \texttt{)}, \texttt{[}, \texttt{]}, \texttt{H}, \texttt{B}, \texttt{C}, \texttt{N}, \texttt{O}, \texttt{F}, \texttt{Si}, \texttt{P}, \texttt{S}, \texttt{Cl}, \texttt{Br}, \texttt{Sn}, \texttt{I}, \texttt{c}, \texttt{n}, \texttt{o}, \texttt{p}, \texttt{s}\}. The minimum, median, and maximum lengths of a SMILES string are 8, 42, and 86, respectively. A special symbol indicating the end of a sequence is appended at the end of each sequence.

\begin{figure}[!t]
    \centering
    \includegraphics[width=\textwidth]{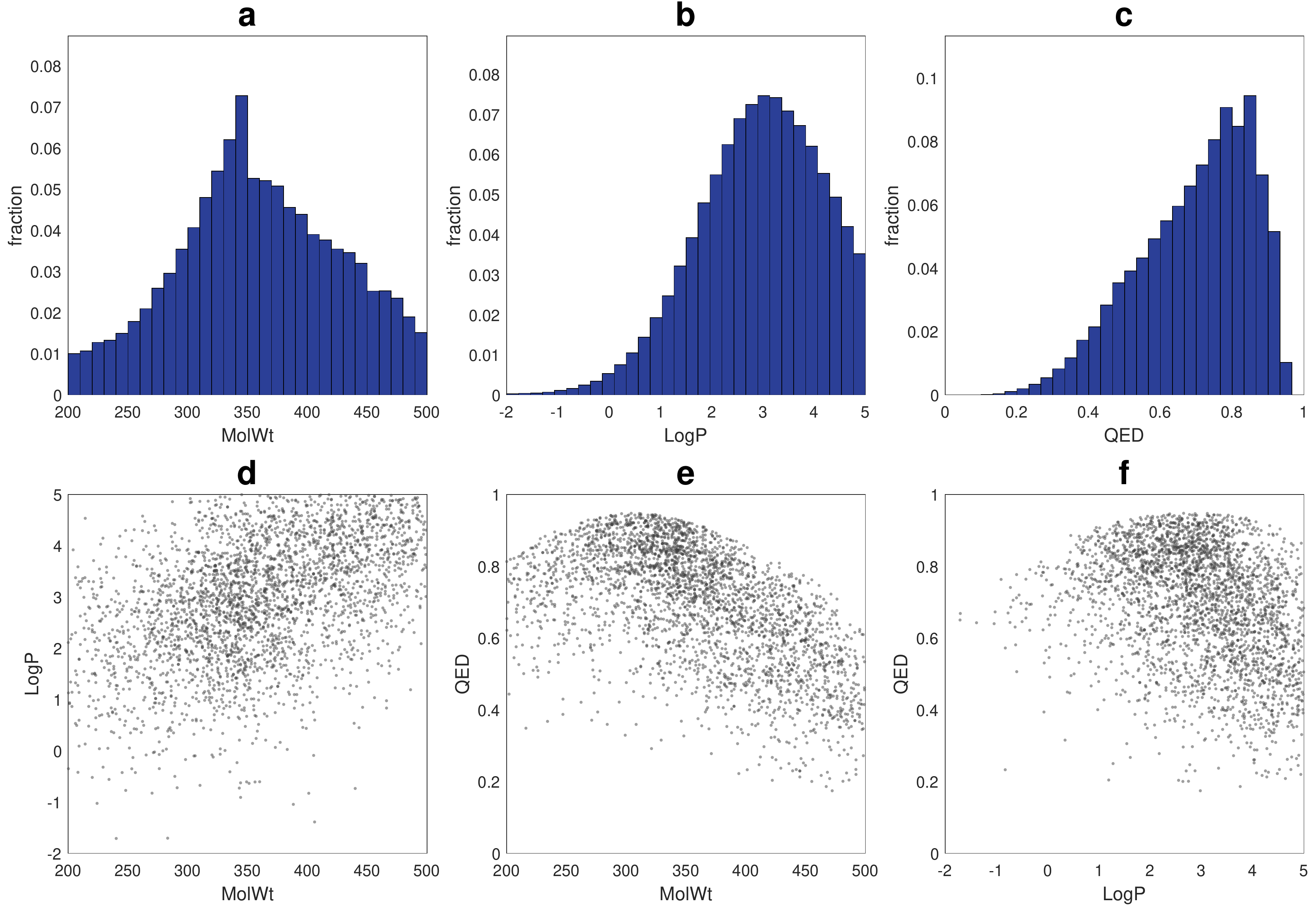}
    \caption{Distribution of properties in training set: histogram of \textbf{(a)} MolWt, \textbf{(b)} LogP, and \textbf{(c)} QED; scatterplot of between \textbf{(d)} MolWt and LogP, \textbf{(e)} MolWt and QED, and \textbf{(f)} LogP and QED.}
    \label{figure:2}
\end{figure}

It is time-consuming and costly to directly obtain the chemical properties of numerous newly generated molecules by performing first-principles calculations or experimental syntheses. In order to efficiently evaluate the proposed approach, we use the three properties that can be readily calculated using the RDKit package:\cite{rdkit} molecular weight (MolWt), Wildman-Crippen partition coefficient (LogP),\cite{LOGP} and quantitative estimation of drug-likeness (QED).\cite{QED} \autoref{figure:2} shows the distributions of these properties in the training set. In this figure, the histograms plot the distributions of individual properties, and the scatterplots represent the pairwise distributions of the properties on 3,000 randomly selected molecules. MolWt ranges from 200 to 500 g/mol and LogP has the range of [-2, 5], according to the drug-like criteria of the ZINC database. QED is valued from 0 to 1 by definition. The averages of MolWt, LogP, and QED are 359.019, 2.911, and 0.696, and their standard deviations are 67.669, 1.179, and 0.158, respectively. There is a positive correlation between MolWt and LogP with the correlation coefficient of 0.434, whereas QED is negatively correlated with both MolWt and LogP with the correlation coefficients of -0.548 and -0.298, respectively.   

\subsection{Experiments}

We evaluate the SSVAE model against baseline models in terms of the prediction performance. We vary the number of labeled molecules (5\%, 10\%, 20\%, and 50\% of the training set) to investigate its effect on property prediction. 95\% of the training set is used for training, while the remaining 5\% is used for early stopping. During training, we normalize each output variable to have a mean of 0 and standard deviation of 1. We use backpropagation with the Adam optimizer.\cite{adam01} We set the default learning rate to 0.001 and use a batch size of 200. Training is terminated if the validation error failed to decrease by 1\% over ten consecutive epochs or the number of epochs reached 300. The property prediction performance is evaluated by mean absolute error (MAE) on the test set.

\begin{figure}[!t]
    \centering
    \includegraphics[width=0.5\textwidth]{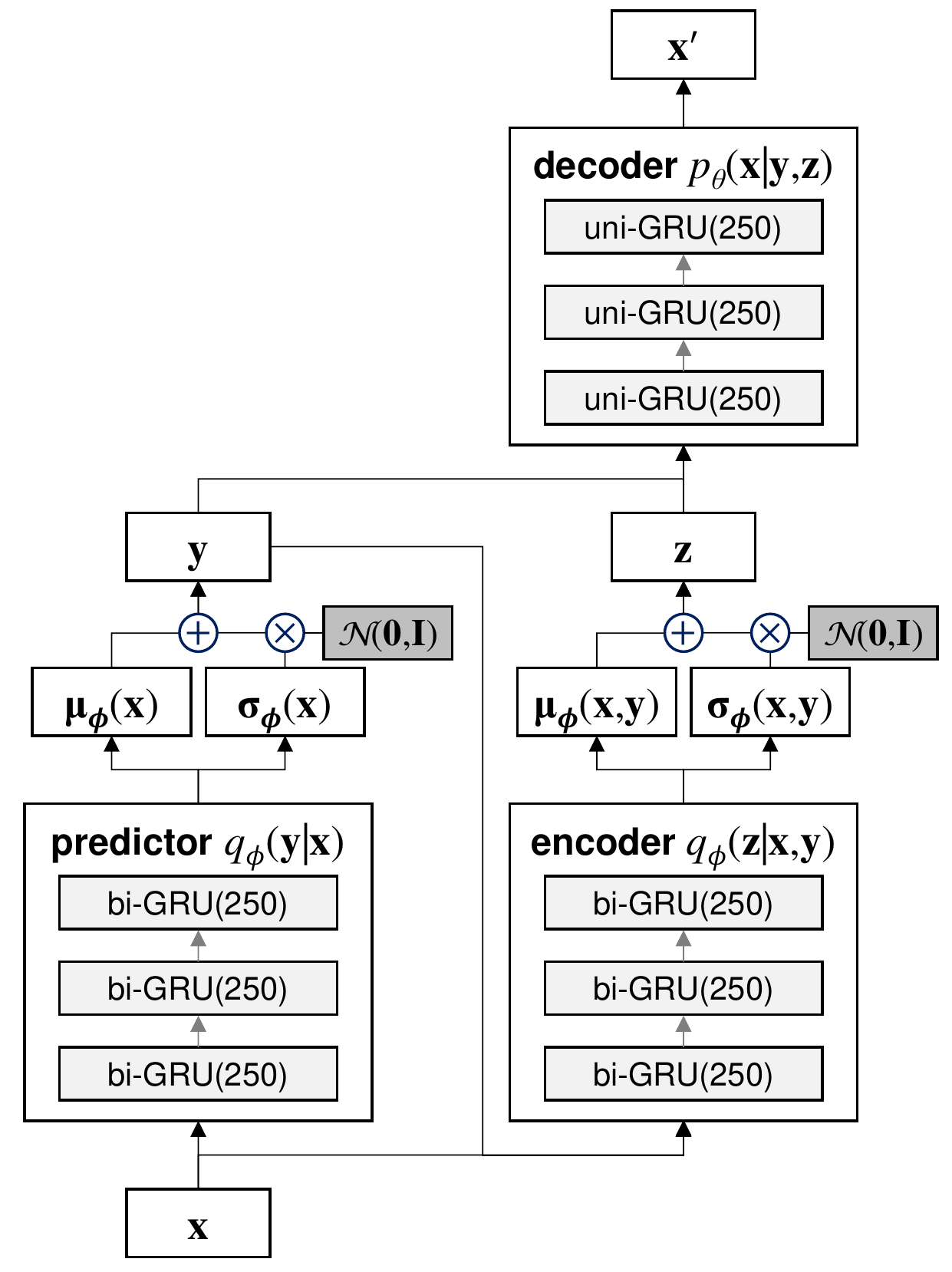}
    \caption{Illustration of SSVAE model architecture.}
    \label{figure:3}
\end{figure}

For the SSVAE model, its predictor, encoder, and decoder networks consist of three hidden layers each having 250 gated recurrent units (GRU).\cite{peek01} The dimension of $\mathbf{z}$ is set as 100. For the prior distribution $p(\mathbf{y})$, we estimate the mean vector ${\widehat{\boldsymbol{\mu}}_\mathbf{y}}$ and covariance matrix $\widehat{\boldsymbol{\Sigma}}_\mathbf{y}$ from the labeled molecules in the training set. \autoref{figure:3} shows the architectural detail of the SSVAE model. We train the model with both the labeled and unlabeled molecules to minimize the objective function in \autoref{eq03}. We set $\beta$ to $10^4$ by conducting a preliminary experiment of minimizing the average validation error of property prediction over the varying numbers of labeled molecules. 

As baseline models, we use the extended-connectivity fingerprint (ECFP),\cite{pred04} molecular graph convolutions (GraphConv),\cite{conv03} independently trained predictor network $q_\phi (\mathbf{y}|\mathbf{x})$, and VAE model jointly trained with a property prediction model that predicts the properties from the latent representation (VAE$_\texttt{property}$),\cite{rnn04} In the cases of the ECFP and GraphConv models, a molecule is processed by three hidden layers, each of which consisting of 2000, 500, and 500 sigmoid units with a dropout rate of 0.2.\cite{dropout} It is then followed by a final output layer that predicts the three output variables. The baseline models except the VAE$_\texttt{property}$ model are trained only with the labeled molecules in the training set to minimize mean squared error between the actual and predicted properties. For the implementation of the VAE$_\texttt{property}$ model\cite{rnn04}, its VAE part is trained with the entire molecules without their labels and the joint prediction model is trained only with the labeled molecules using the same objective function as that of the SSVAE model.

To demonstrate conditional molecular design, we use the SSVAE model trained on the training set in which 50\% of the molecules were labeled with their properties. New molecules are generated under various target conditions of properties, each of which sets one property with a specific target value and the others to be sampled from the corresponding conditional prior distribution.

We compare the SSVAE model with the unsupervised VAE model (VAE$_\texttt{unsupervised}$) and the VAE$_\texttt{property}$ model that are trained on the same training set. In the case of the VAE$_\texttt{property}$ model, we use Gaussian process to smoothly approximate the property prediction given a latent representation and perform Bayesian optimization.\cite{rnn04} The objective function for Bayesian optimization is set as the normalized absolute difference between the target value and the value predicted from the latent representation by the joint property prediction model. For generating a molecule, Bayesian optimization is terminated when the value of the objective function is below 0.01.

Molecules are generated from the decoder network of the model using beam search, where the beam width $K$ is set to 5. The target values for MolWt, LogP, and QED are set as \{250, 350, 450\}, \{1.5, 3.0, 4.5\}, and \{0.5, 0.7, 0.9\}, respectively. We also test generating new molecules unconditionally without specifying any target value. During the generation procedure given each target condition, we check the validity of each generated molecule using the SMILES grammar rules (\emph{e.g.}, the number of open/close parentheses and the existence of unclosed rings) and pre-conditions (\emph{e.g.}, kekulizability) using the RDKit package.\cite{rdkit} We discard those molecules that are identified as invalid, already exist in the training set, or duplicated. The generation procedure continues until 3,000 novel unique molecules are obtained or the number of trials exceeds 10,000. Then, these molecules are labeled with MolWt, LogP and QED to confirm whether their properties were distributed around their respective target values.

All the experiments are implemented based on GPU-accelerated TensorFlow in Python.\cite{tensorflow} The source code of the SSVAE model used in the experiments is available at \url{https://github.com/nyu-dl/conditional-molecular-design-ssvae}.

\subsection{Property Prediction}

\autoref{tab:1} shows the results in terms of MAE with the varying fractions of labeled molecules in the training set. We report the average and standard deviation over ten repetitions for each setting. Among the baseline models, the GraphConv model was superior to the ECFP model in every case. The GraphConv model yielded performance comparable to the predictor model with a fewer labeled molecules, while the predictor model was superior with more labeled molecules. The predictor model significantly outperformed the ECFP and GraphConv models on predicting MolWt, which is almost identical to the task of simply counting atoms in a SMILES string. The VAE$_\texttt{property}$ model performed worse for MolWt but was superior in predicting LogP and QED with a fewer labeled molecules, when compared to the predictor model.

The SSVAE model outperformed the baseline models on most of the cases. The SSVAE model yielded better prediction performance than the predictor model did with a lower fraction of labeled molecules. On the other hand, the difference between the SSVAE model and the predictor model narrowed as the fraction of labeled molecules increased. The results successfully demonstrate the effectiveness of this semi-supervised learning scheme in improving property prediction. 

\begin{table}[!t]	
	\caption{Property prediction performance with varying fractions of labeled molecules.}
	\label{tab:1}
	\centering
    \small
	\scalebox{0.75}{
	\begin{tabular}{ll|rrcrr}
		\hline
		\thead{frac. labeled} & \thead{property} & \thead{ECFP} & \thead{GraphConv} & \thead{predictor network} & \thead{VAE$_\texttt{property}$} & \thead{SSVAE} \\
		\hline
        5\% & MolWt &17.713$\pm$0.396&6.723$\pm$2.116&2.582$\pm$0.288&  3.463$\pm$0.971&1.639$\pm$0.577\\
        & LogP &0.380$\pm$0.009&0.187$\pm$0.015&0.162$\pm$0.006& 0.125$\pm$0.013&0.120$\pm$0.006\\
        & QED &0.053$\pm$0.001&0.034$\pm$0.004&0.037$\pm$0.002&  0.029$\pm$0.002&0.028$\pm$0.001\\
        \hline
        10\% & MolWt &15.057$\pm$0.358&5.255$\pm$0.767&1.986$\pm$0.470& 2.464$\pm$0.581&1.444$\pm$0.618\\
        & LogP &0.335$\pm$0.005&0.148$\pm$0.016&0.116$\pm$0.006& 0.097$\pm$0.008&0.090$\pm$0.004\\
        & QED &0.045$\pm$0.001&0.028$\pm$0.003&0.027$\pm$0.002& 0.021$\pm$0.002&0.021$\pm$0.001\\
        \hline
        20\% & MolWt &12.047$\pm$0.168&4.597$\pm$0.419&1.228$\pm$0.229& 1.748$\pm$0.266 &1.008$\pm$0.370\\
        & LogP &0.249$\pm$0.004&0.112$\pm$0.015&0.070$\pm$0.007& 0.074$\pm$0.006 &0.071$\pm$0.007\\
        & QED &0.033$\pm$0.001&0.021$\pm$0.002&0.017$\pm$0.002& 0.015$\pm$0.001 &0.016$\pm$0.001\\
        \hline
        50\% & MolWt &9.012$\pm$0.184&4.506$\pm$0.279&1.010$\pm$0.250& 1.350$\pm$0.319 &1.050$\pm$0.164\\
        & LogP &0.180$\pm$0.003&0.086$\pm$0.012&0.045$\pm$0.005& 0.049$\pm$0.008 &0.047$\pm$0.003\\
        & QED &0.023$\pm$0.000&0.018$\pm$0.001&0.011$\pm$0.001& 0.009$\pm$0.002 &0.010$\pm$0.001\\
		\hline
	\end{tabular}
	}
\end{table}

\subsection{Conditional Molecular Design}

\autoref{tab:2} shows the statistics of generated molecules given each target condition in order to investigate the efficacy of molecule generation. For the SSVAE model, the fraction of invalid molecules was generally less than 1\%, and was slightly higher when the target value had a lower density in the distribution of the training set. There were a few duplicated molecules from unconditional generation, and the fraction of new unique molecules was 92.7\%. There were more duplicates when molecules were conditionally generated. In particular, the fraction of duplicated molecules for a target condition was higher when the prediction of the property for the condition was more accurate. As the normalized MAEs of MolWt, LogP, and QED by the SSVAE model were 0.016, 0.038, and 0.058, MolWt yielded the lowest fraction of new unique molecules and was followed by LogP and QED.

Both the VAE$_\texttt{unsupervised}$ and VAE$_\texttt{property}$ models were less efficient than the SSVAE model was, evident from the higher number of duplicated molecules generated. When we tried sampling from the VAE$_\texttt{unsupervised}$ model without beam search, the model rarely generated duplicated ones, while the majority of the generated ones were invalid.

\begin{table}[!t]	
	\caption{Molecule generation efficacy of conditional molecular design.}
	\label{tab:2}
	\centering
    \small 
	\scalebox{0.75}{
	\begin{tabular}{ll|rrrrr}
		\hline
		\thead{model} & \thead{target condition} & \thead{no. generated} & \thead{no. invalid} & \thead{no. in training set} & \thead{no. duplicated} & \thead{no. new unique}\\
        \hline
        VAE$_\texttt{unsupervised}$ &uncond. gen. (sampling) & 10000 (100\%) & 8771 (87.7\%) & 5 (0.1\%) & 65 (0.7\%) & 1159 (11.6\%) \\
        &uncond. gen. (beam search) & 10000 (100\%) & 2 (0.0\%) & 1243 (12.4\%) & 6802 (68.0\%) & 1953 (19.5\%) \\
        \hline
        {VAE$_\texttt{property}$} &unconditional generation & 5940 (100\%) & 2 (0.0\%) & 486 (8.2\%) & 2452 (41.3\%) & 3000 (50.5\%) \\
        \cline{2-7}
        &MolWt=250 & 10000 (100\%)  & 7 (0.1\%)& 1136 (11.4\%)& 6400 (64.0\%)& 2457 (24.6\%)\\
        &MolWt=350 & 8618 (100\%) & 1 (0.0\%)& 647 (7.5\%)& 4970 (57.7\%)& 3000 (34.8\%) \\
        &MolWt=450 & 10000 (100\%) & 9 (0.1\%)& 1120 (11.2\%) & 6626 (66.3\%)& 2245 (22.5\%)\\
        \cline{2-7}
        &LogP=1.5 & 9521 (100\%)  & 10 (0.1\%)& 575 (6.0\%)& 5936 (62.3\%)& 3000 (31.5\%)\\
        &LogP=3.0 & 7628 (100\%)& 4 (0.1\%)& 560 (7.3\%)& 4064 (53.3\%)& 3000 (39.3\%)\\
        &LogP=4.5 & 10000 (100\%)& 13 (0.1\%)& 862 (8.6\%)& 6563 (65.6\%)& 2562 (25.6\%)\\
        \cline{2-7}
        &QED=0.5 & 9643 (100\%) & 20 (0.2\%)& 764 (7.9\%)& 5859 (60.8\%)& 3000 (31.1\%) \\
        &QED=0.7 & 6888 (100\%)& 3 (0.0\%)& 617 (9.0\%)& 3268 (47.4\%)& 3000 (43.6\%)\\
        &QED=0.9 & 10000 (100\%)& 6 (0.1\%)& 851 (8.5\%)& 6476 (64.8\%)& 2667 (26.7\%)\\
        \hline
        SSVAE&unconditional generation &3236 (100\%)&23 (0.7\%)&163 (5.0\%)&50 (1.5\%)&3000 (92.7\%)\\
        \cline{2-7}
        &MolWt=250 &4079 (100\%)&16 (0.4\%)&177 (4.3\%)&886 (21.7\%)&3000 (73.5\%)\\
        &MolWt=350 &3629 (100\%)&17 (0.5\%)&137 (3.8\%)&475 (13.1\%)&3000 (82.7\%)\\
        &MolWt=450 &4181 (100\%)&31 (0.7\%)&277 (6.6\%)&873 (20.9\%)&3000 (71.8\%)\\
        \cline{2-7}
        &LogP=1.5 &3457 (100\%)&26 (0.8\%)&127 (3.7\%)&304 (8.8\%)&3000 (86.8\%)\\
        &LogP=3.0 &3433 (100\%)&21 (0.6\%)&166 (4.8\%)&246 (7.2\%)&3000 (87.4\%)\\
        &LogP=4.5 &3507 (100\%)&30 (0.9\%)&186 (5.3\%)&291 (8.3\%)&3000 (85.5\%)\\
        \cline{2-7}
        &QED=0.5 & 3456 (100\%)&49 (1.4\%)&171 (4.9\%)&236 (6.8\%)&3000 (86.8\%)\\
        &QED=0.7 & 3308 (100\%)&19 (0.6\%)&168 (5.1\%)&121 (3.7\%)&3000 (90.7\%)\\
        &QED=0.9 & 3233 (100\%)&12 (0.4\%)&125 (3.9\%)&96 (3.0\%)&3000 (92.8\%)\\
        \hline
	\end{tabular}
	}
\end{table}

\begin{table}[!t]	
	\caption{Comparison between original training set and conditional molecular design results.}
	\label{tab:3}
	\centering
    \small 
	\scalebox{0.75}{
	\begin{tabular}{ll|rcccc}
		\hline
		\thead{model} & \thead{target condition} & \thead{no. molecules} & \thead{sequence length} & \thead{MolWt} & \thead{LogP} & \thead{QED}\\
		\hline
        training set & all molecules & 300000 & 42.375$\pm$9.316& 359.019$\pm$67.669 & 2.911$\pm$1.179 & 0.696$\pm$0.158\\ 
        & labeled molecules & 150000&42.402$\pm$9.313&359.381$\pm$67.666&2.912$\pm$1.177&0.696$\pm$0.158\\
        \cline{2-7}
        & 240$\leq$MolWt$\leq$260 & 4868&28.818$\pm$3.707&250.237$\pm$5.642&2.116$\pm$1.074&0.762$\pm$0.118\\
        & 340$\leq$MolWt$\leq$360 & 18799&41.094$\pm$3.865&348.836$\pm$5.751&2.793$\pm$1.094&0.759$\pm$0.123\\
        & 440$\leq$MolWt$\leq$460 & 8546&53.562$\pm$4.586&448.959$\pm$5.631&3.569$\pm$0.989&0.540$\pm$0.122\\
        \cline{2-7}
        & 1.4$\leq$LogP$\leq$1.6 & 4591&38.223$\pm$8.679&320.299$\pm$61.256&1.503$\pm$0.057&0.757$\pm$0.130\\
        & 2.9$\leq$LogP$\leq$3.1 & 9657&42.214$\pm$9.044&357.686$\pm$62.747&2.999$\pm$0.058&0.722$\pm$0.150\\
        & 4.4$\leq$LogP$\leq$4.6 & 6040&47.228$\pm$8.404&404.425$\pm$56.184&4.496$\pm$0.059&0.589$\pm$0.149\\
        \cline{2-7}
        & 0.49$\leq$QED$\leq$0.51 & 3336&49.028$\pm$8.415&409.733$\pm$63.362&3.467$\pm$1.130&0.500$\pm$0.006\\
        & 0.69$\leq$QED$\leq$0.71 & 5961&42.614$\pm$8.571&362.448$\pm$63.359&2.929$\pm$1.255&0.700$\pm$0.006\\
        & 0.89$\leq$QED$\leq$0.91 & 5466&36.910$\pm$5.219&316.128$\pm$32.789&2.529$\pm$0.865&0.900$\pm$0.006\\
        \hline
        VAE$_\texttt{unsupervised}$ & uncond. gen. (sampling) & 1159 & 34.965$\pm$8.123& 305.756$\pm$65.817& 2.900$\pm$1.262 &0.725$\pm$0.143 \\					
        & uncond. gen. (beam search) & 1953 & 43.911$\pm$8.384& 366.502$\pm$60.865& 2.987$\pm$0.995 & 0.707$\pm$0.149\\ 
        \hline
        {VAE$_\texttt{property}$} & unconditional generation & 3000&43.853$\pm$7.477&362.037$\pm$54.528&2.986$\pm$0.994&0.716$\pm$0.132\\
        \cline{2-7}
        & MolWt=250 & 2457&30.284$\pm$4.261&255.676$\pm$25.457&2.230$\pm$0.965&0.789$\pm$0.093\\
        & MolWt=350 & 3000&40.718$\pm$4.534&338.858$\pm$27.478&3.023$\pm$0.982&0.766$\pm$0.111\\
        & MolWt=450 & 2245&53.738$\pm$4.636&443.253$\pm$23.950&3.477$\pm$0.946&0.573$\pm$0.108\\
        \cline{2-7}
        & LogP=1.5 & 3000&40.296$\pm$8.019&329.754$\pm$58.057&1.478$\pm$0.537&0.744$\pm$0.117\\
        & LogP=3.0 & 3000&42.975$\pm$7.352&353.535$\pm$53.487&2.740$\pm$0.643&0.728$\pm$0.127\\
        & LogP=4.5 & 2562&46.198$\pm$7.871&389.465$\pm$53.514&4.290$\pm$0.435&0.636$\pm$0.136\\
        \cline{2-7}
        & QED=0.5 & 3000&49.955$\pm$6.749&409.021$\pm$47.190&3.386$\pm$1.023&0.544$\pm$0.096\\
        & QED=0.7 & 3000&45.331$\pm$7.382&375.083$\pm$53.888&3.079$\pm$1.002&0.688$\pm$0.111\\
        & QED=0.9 & 2667&37.441$\pm$5.573&310.396$\pm$38.871&2.515$\pm$0.918&0.860$\pm$0.062\\
        \hline
        {SSVAE} & unconditional generation & 3000 & 42.093$\pm$9.010&359.135$\pm$65.534&2.873$\pm$1.117&0.695$\pm$0.148\\
        \cline{2-7}
        & MolWt=250 & 3000 & 28.513$\pm$3.431&250.287$\pm$6.742&2.077$\pm$1.072&0.796$\pm$0.094\\
        & MolWt=350 & 3000 & 41.401$\pm$4.393&349.599$\pm$7.345&2.782$\pm$1.060&0.723$\pm$0.129\\
        & MolWt=450 & 3000 & 53.179$\pm$4.760&449.593$\pm$8.901&3.544$\pm$1.016&0.563$\pm$0.122\\
        \cline{2-7}
        & LogP=1.5 & 3000 & 38.709$\pm$8.669&323.336$\pm$60.288&1.539$\pm$0.301&0.750$\pm$0.127\\
        & LogP=3.0 & 3000 & 42.523$\pm$8.919&361.264$\pm$61.524&2.984$\pm$0.295&0.701$\pm$0.149\\
        & LogP=4.5 & 3000 & 45.566$\pm$8.698&397.609$\pm$61.436&4.350$\pm$0.309&0.624$\pm$0.147\\
        \cline{2-7}
        & QED=0.5 & 3000 & 48.412$\pm$7.904&404.159$\pm$56.788&3.288$\pm$1.069&0.527$\pm$0.094\\
        & QED=0.7 & 3000 & 41.737$\pm$7.659&356.672$\pm$55.629&2.893$\pm$1.093&0.719$\pm$0.088\\
        & QED=0.9 & 3000 & 36.243$\pm$7.689&312.985$\pm$56.270&2.444$\pm$1.079&0.840$\pm$0.070\\
        \hline
	\end{tabular}
	}
\end{table}

\begin{figure}[!t]
    \centering
    \includegraphics[width=\textwidth]{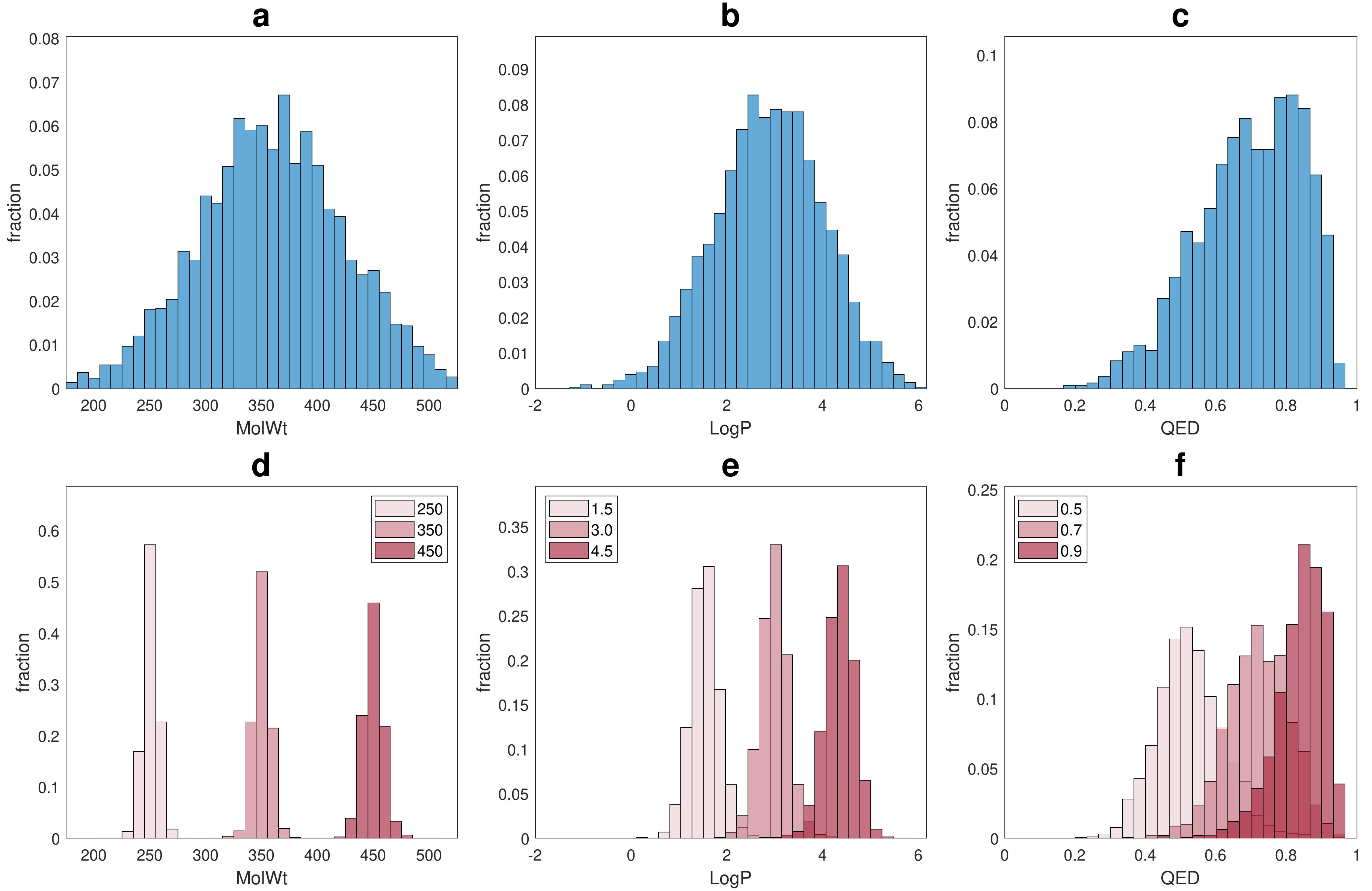}
    \caption{Distribution of properties by conditional molecular design using SSVAE model (proposed): histogram of unconditional generation results for \textbf{(a)} MolWt, \textbf{(b)} LogP, and \textbf{(c)} QED; histogram of conditional generation results for \textbf{(d)} MolWt, \textbf{(e)} LogP, and \textbf{(f)} QED.}
    \label{figure:4}
\end{figure}

\begin{figure}[!t]
    \centering
    \includegraphics[width=\textwidth]{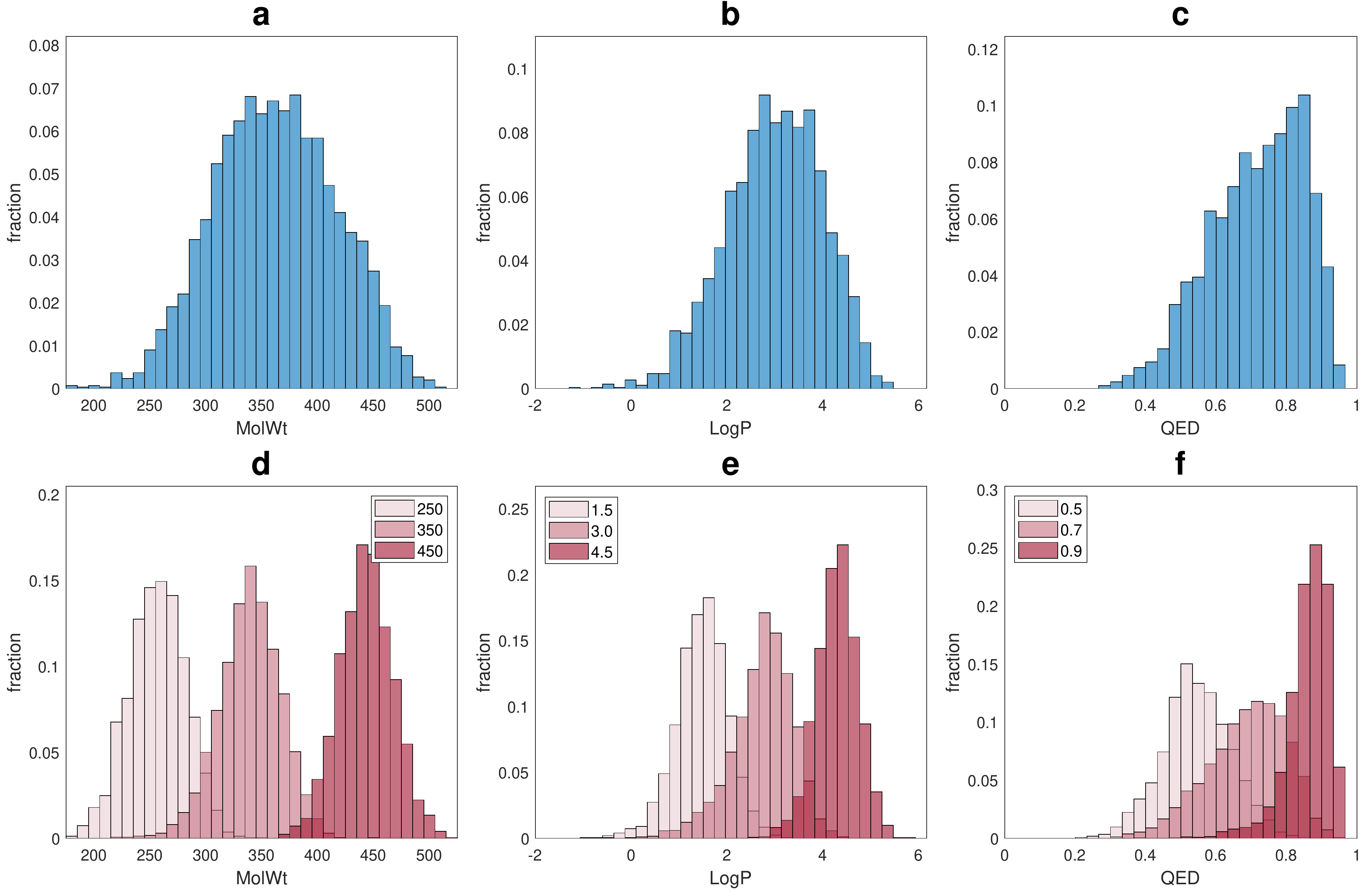}
    \caption{Distribution of properties by conditional molecular design using VAE$_\texttt{property}$ model (baseline): histogram of unconditional generation results for \textbf{(a)} MolWt, \textbf{(b)} LogP, and \textbf{(c)} QED; histogram of conditional generation results for \textbf{(d)} MolWt, \textbf{(e)} LogP, and \textbf{(f)} QED.}
    \label{figure:5}
\end{figure}

\autoref{tab:3} presents the summary statistics for newly generated molecules of each condition, and \autoref{figure:4} and \ref{figure:5} compare the histograms representing the distributions of MolWt, LogP, and QED between different target conditions by the SSVAE and VAE$_\texttt{property}$ models, respectively. For the SSVAE model, the unconditionally generated molecules without any target value followed the property distributions of the training set, as evident from contrasting \autoref{figure:4}\textbf{(a--c)} and \autoref{figure:2}\textbf{(a--c)}. When a target condition is set, the SSVAE model successfully generated new molecules fulfilling the condition. In \autoref{figure:4}\textbf{(d--f)}, we observe that the distributions of the conditionally generated molecules by the SSVAE model were centered around the corresponding target values with much smaller standard deviations. The conditionally generated molecules followed the property distributions of those molecules in the training set whose property was around the target value, as shown in \autoref{tab:3}. The accuracy of conditional molecular design for a target condition tended to be proportional to the prediction accuracy of the corresponding property. For MolWt which yielded the lowest normalized MAE, the distributions were relatively narrow and separated distinctly by its target values. On the other hand, LogP and QED exhibited larger overlap between target values. The VAE$_\texttt{property}$ model also generated new molecules satisfying the target conditions, but the distributions were relatively dispersed and far from the corresponding target values compared to those by the SSVAE model, as shown in \autoref{figure:5}\textbf{(d--f)}.

\begin{figure}[!t]
    \centering
    \includegraphics[width=\textwidth]{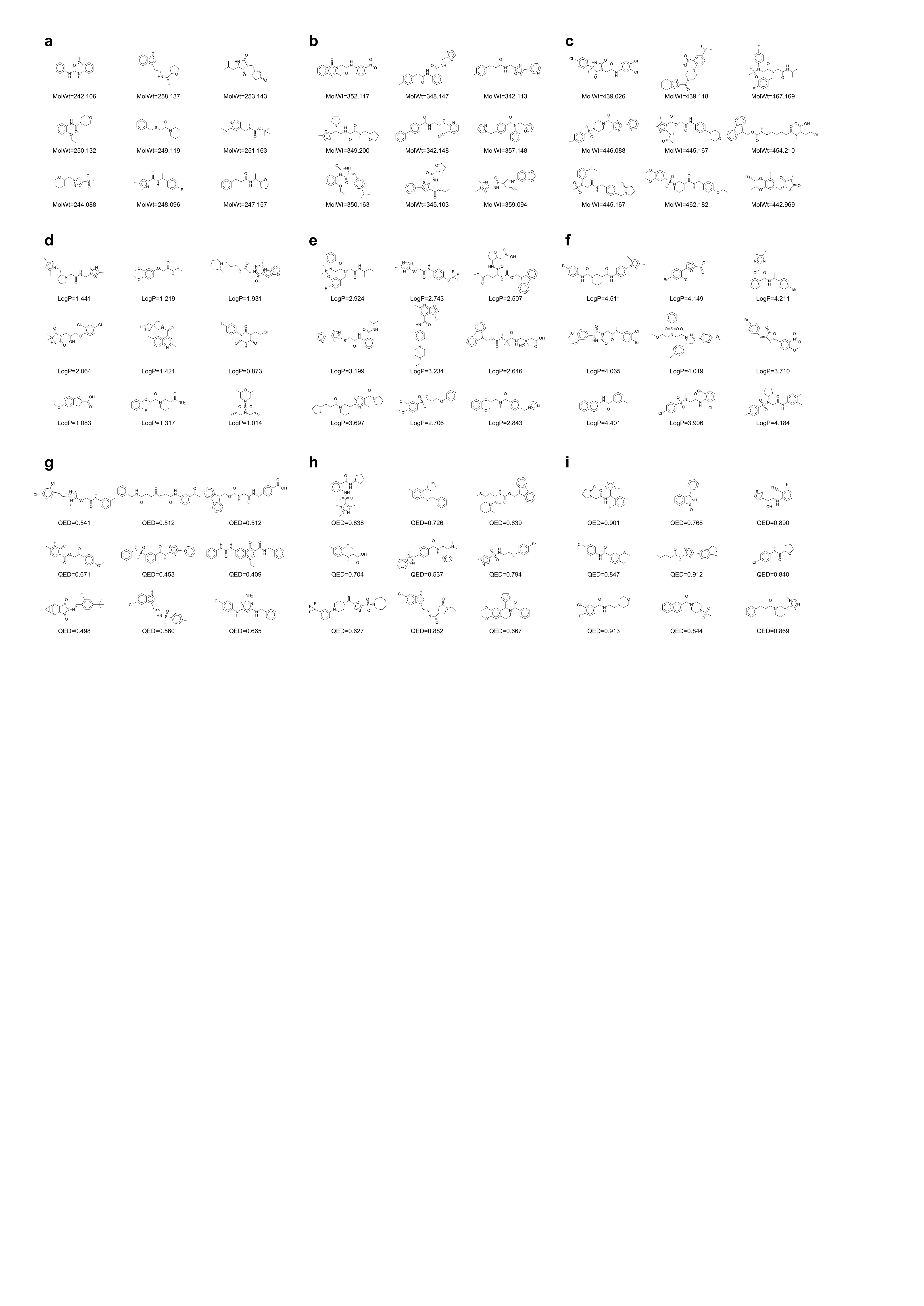}
    \caption{Examples of generated molecules using SSVAE model conditioning on \textbf{(a)} MolWt=250, \textbf{(b)} MolWt=350, \textbf{(c)} MolWt=450, \textbf{(d)} LogP=1.5, \textbf{(e)} LogP=3.0, \textbf{(f)} LogP=4.5, \textbf{(g)} QED=0.5, \textbf{(h)} QED=0.7, and \textbf{(i)} QED=0.9.}
    \label{figure:6}
\end{figure}

We present some sample molecules generated from the SSVAE model under each target condition in \autoref{figure:6}. From the glance at the sample molecules generated with three different target MolWt, we observe that the SSVAE had generated smaller molecules when the target condition of MolWT was set to 250. On the other hand, when MolWt was set to a higher value, relatively larger molecules were generated.

\begin{table}[!t]	
	\caption{Training and inference time comparison.}
	\label{tab:4}
	\centering
    \small
	\begin{tabular}{l|rrr}
		\hline
		\thead{model} & \thead{training time} & \multicolumn{2}{c}{\thead{inference time (per generation)}} \\
        \cline{3-4}
        & & \thead{unconditional gen.} & \thead{conditional gen.} \\
		\hline
VAE$_\texttt{unsupervised}$ & 7.4$\pm$1.9 hrs & 4.9$\pm$0.7 s & -\\
{VAE$_\texttt{property}$} & 10.2$\pm$2.1 hrs & 4.7$\pm$0.7 s & 46.6$\pm$135.5 s\\
SSVAE & 20.3$\pm$5.3 hrs & 4.6$\pm$1.0 s & 4.5$\pm$1.1 s\\
		\hline
	\end{tabular}
\end{table}

\autoref{tab:4} compares training and inference time between the models. It took longer to train the SSVAE model than the other models, because it has one more RNN as the predictor network compared to the other models. For unconditional generation, there was a slight difference in the generation speed between the models. Conditional generation with the VAE$_\texttt{property}$ model, which involves Bayesian optimization, was time-consuming. On the other hand, conditional generation with the SSVAE model, which simply uses the decoder network without any extra optimization procedure, was as fast as unconditional generation.

\section{Conclusion}

We have presented a novel approach to conditionally generating molecules efficiently and accurately using the regression version of SSVAE. We designed and trained the SSVAE model on a partially labeled training set in which only a small portion of molecules were labeled with their properties. New molecules with desired properties were generated from the generative distribution of the SSVAE model given a target condition of properties. The experiments using drug-like molecules sampled from the ZINC database have successfully demonstrated the effectiveness in terms of both property prediction and conditional molecular design. The SSVAE model efficiently generates novel molecules satisfying the target conditions without any extra optimization procedure. Moreover, the conditional design procedure works by automatically learning implicit knowledge from data without necessitating any explicit knowledge.

The proposed approach can serve as an efficient tool for designing new chemical structures fulfilling a specified target condition. These structures generated as SMILES strings are to be examined further to obtain realistic molecules with desired properties. In this study, the application is limited to only a part of the chemical space that SMILES can represent. To broaden its applicability, we should investigate other alternatives to SMILES that provide higher coverage of the chemical space and are able to represent molecules more comprehensively.

\begin{acknowledgement}
This work was supported by the National Research Foundation of Korea (NRF) grant funded by the Korea government (MSIT; Ministry of Science and ICT) (No. NRF-2017R1C1B5075685). K.C. thanks support by AdeptMind, eBay, TenCent, NVIDIA and CIFAR. K.C. was partly supported for this work by Samsung Electronics (Improving Deep Learning using Latent Structure).
\end{acknowledgement}

\section{Conflict of Interest}
The authors declare no competing financial interest.

\bibliography{refs}

\end{document}